\documentclass[runningheads]{llncs}
\usepackage[T1]{fontenc}

\usepackage{graphicx}
\usepackage{hyperref}
\usepackage{amsfonts}
\usepackage{csquotes}
\usepackage{amsmath} 
\usepackage{cleveref}
\usepackage{booktabs}
\usepackage{multirow}
\usepackage{booktabs}
\usepackage{pifont}
\usepackage{url}   
\usepackage{wrapfig}
\usepackage{marvosym}
\usepackage{bbm}
\newcommand{\mypara}[1]{%
  \par\addvspace{0.6\baselineskip}%
  \noindent\textbf{#1}\hspace{0.6em}\ignorespaces%
}

\makeatletter
\newcommand{\maketitlesupplementary}{%
  \clearpage
  \setcounter{page}{1}
  \setcounter{section}{0}
  \setcounter{subsection}{0}
  \setcounter{figure}{0}
  \setcounter{table}{0}

  \renewcommand{\thesection}{S\arabic{section}}
  \renewcommand{\thesubsection}{S\arabic{section}.\arabic{subsection}}
  \renewcommand{\thefigure}{S\arabic{figure}}
  \renewcommand{\thetable}{S\arabic{table}}

  \providecommand{\theHsection}{}
  \providecommand{\theHsubsection}{}
  \providecommand{\theHfigure}{}
  \providecommand{\theHtable}{}
  \providecommand{\theHpage}{}
  \renewcommand{\theHsection}{S.\arabic{section}}
  \renewcommand{\theHsubsection}{S.\arabic{section}.\arabic{subsection}}
  \renewcommand{\theHfigure}{S.\arabic{figure}}
  \renewcommand{\theHtable}{S.\arabic{table}}
  \renewcommand{\theHpage}{S.\arabic{page}}

  \section*{Supplementary Material}
}
\makeatother

\begin{document}
\title{OccFace: Unified Occlusion-Aware Facial Landmark Detection with Per-Point Visibility}

%
\titlerunning{OccFace: Unified Occlusion-Aware Facial Landmark Detection}
%
\author{Xinhao Xiang\inst{1}\thanks{This work was done when Xinhao Xiang was an intern at Genies inc.}
\and
Zhengxin Li\inst{2}
\and
Saurav Dhakad\inst{2}
\and
Theo Bancroft\inst{3}
\and
Jiawei Zhang\inst{1}
\and
Weiyang Li\inst{2}
}

\authorrunning{X. Xiang et al.}
%
\institute{
IFM Lab, University of California, Davis, CA, USA\\
\email{\{xhxiang,jiwzhang\}@ucdavis.edu}
\and
Genies inc., CA, USA \\
\email{wli@genies.com}
\and
University of Arizona, AZ, USA 
}
\maketitle              

\begin{abstract}

Accurate facial landmark detection under occlusion remains challenging, especially for human-like faces with large appearance variation and rotation-driven self-occlusion.
Existing detectors typically localize landmarks while handling occlusion implicitly, without predicting per-point visibility that downstream applications can benefits.
We present OccFace, an occlusion-aware framework for universal human-like faces, including humans, stylized characters, and other non-human designs.
OccFace adopts a unified dense 100-point layout and a heatmap-based backbone, and adds an occlusion module that jointly predicts landmark coordinates and per-point visibility by combining local evidence with cross-landmark context.
Visibility supervision mixes manual labels with landmark-aware masking that derives pseudo visibility from mask--heatmap overlap.
We also create an occlusion-aware evaluation suite reporting NME on visible vs.\ occluded landmarks and benchmarking visibility with Occ AP, F1@0.5, and ROC-AUC, together with a dataset annotated with 100-point landmarks and per-point visibility.
Experiments show improved robustness under external occlusion and large head rotations, especially on occluded regions, while preserving accuracy on visible landmarks.

\keywords{Facial landmark detection \and Cross-domain face alignment \and  Visibility estimation.}
\end{abstract}    
\section{Introduction}
\label{sec:intro}

\begin{figure}[!t]
  \centering
  \includegraphics[width=0.83\linewidth]{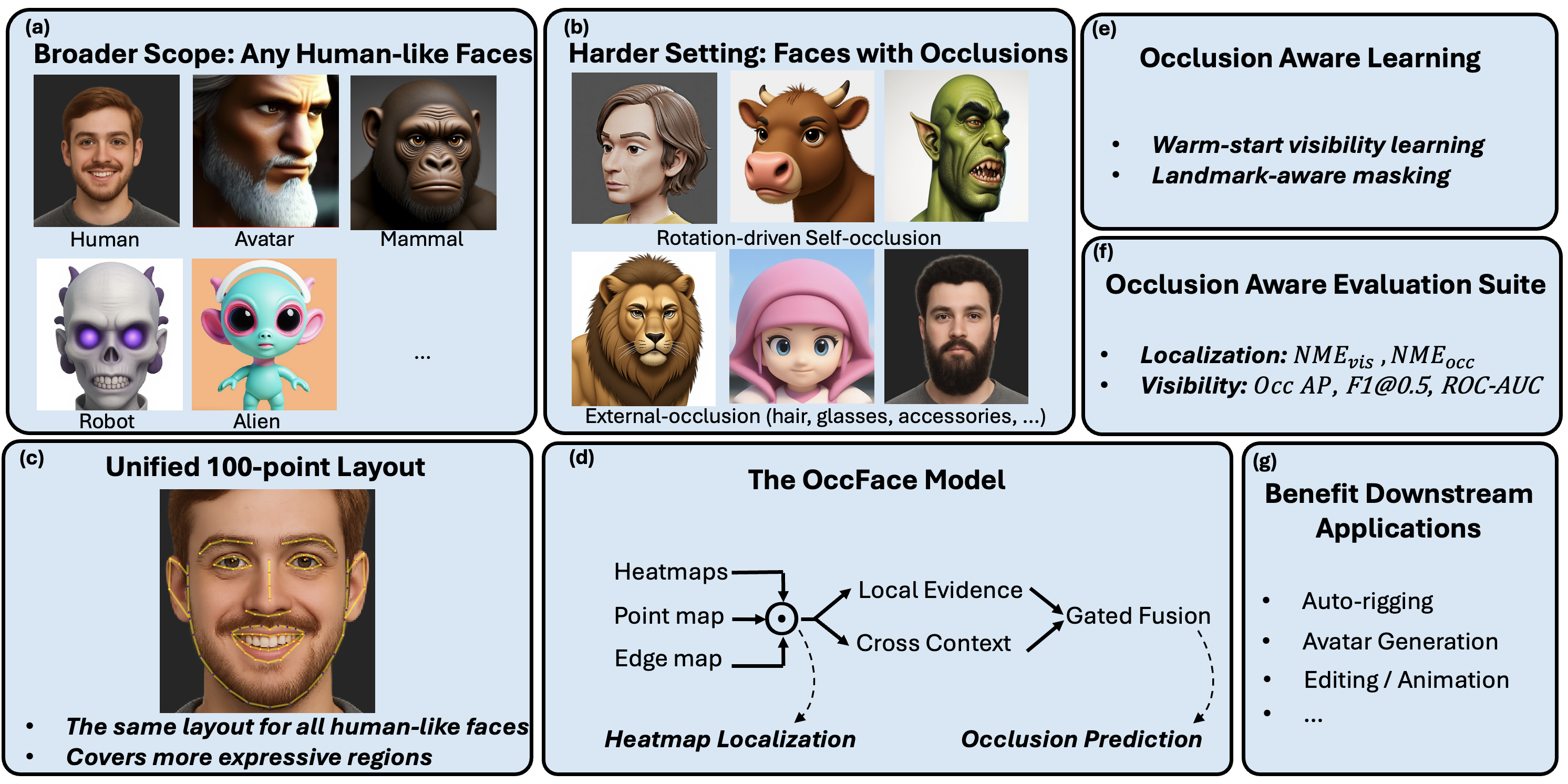}
  \caption{
(a) We study universal human-like face alignment beyond real humans, covering diverse domains.
(b) The setting is harder under occlusions, including both external occludes and rotation-driven self-occlusion under large head pose.
(c) To make landmarks consistent across such domains, we introduce a unified 100-point layout that adds expressive structures (e.g., ear contours, inner-mouth, fine-grained eyes).
(d) We design \textbf{OccFace} model to keep the strong spatial inductive bias of heatmap localization, while explicitly predicting per-point visibility: auxiliary point/edge evidence maps provide geometric cues that gate heatmap localization and support two visibility cues—local evidence (for external occlusion) and cross-landmark context (for correlated self-occlusion)—combined by a gated fusion module.
(e)-(f) We stabilize visibility learning by proposing additional occlusion-aware training objective and evaluation suite.
(g) The predicted visibility could benefits various downstream face-centric applications.}
  \label{fig:teaser}
\end{figure}

Facial landmarks offer a structured geometric description of the face and play a central role in perception and graphics tasks such as facial animation, geometric reconstruction, retargeting, and dense correspondence analysis~\cite{kazemi2014one}. As these tasks depend on consistent and anatomically meaningful reference points, accurate landmark localization is essential for achieving stable and interpretable downstream results.
While landmark detection has progressed rapidly on realistic human faces~\cite{kumar2020luvli,ADNet,chiang2025orformer}, modern applications increasingly operate on a far wider range of \emph{human-like faces}, ranging from stylized avatars, cartoon and game characters, anthropomorphic animals, and robot-inspired designs~\cite{stricker2018manga,cai2021caricature,cheng2024stylizedfacepoint}. Compared to standard human-face datasets, these domains exhibit substantially larger variation in facial geometry, proportions, appearance, and rendering style.
Such diversity increases the ambiguity and complexity of geometric interpretation, exposing failure modes that are rarely captured conventional existing benchmarks.


Two challenges become especially apparent in this universal setting.
\textit{(1) Layout.} Most existing detectors adopt sparse, human-centric landmark layouts~\cite{COFW,300W} that are tied to adult human morphology and under-specify structures that vary widely across other human-like faces. 
For instance, ears, inner-mouth structure, and fine-grained eye details are often critical for expressing shape and emotion in stylized characters and mammals.
However, current human-centric datasets are absent or weakly represented in their 29/68/98-point layout~\cite{COFW,WFLW,300W}. 
We therefore introduce a denser \textbf{unified 100-point layout} to provide a more expressive and consistent representation across domains.
\textit{(2) Occlusion.} Occlusion becomes significantly more complex when large head rotations are common. In addition to external occludes (such as hair, hands, accessories, fur), head pose induces strong self-occlusion where far-side structures disappear in correlated patterns. Such cases are frequent in practice but are weakly represented in many classic benchmarks~\cite{300W,WFLW}, where their yaw angles are typically limited to only a few degrees, while many real applications regularly encounter rotations of 30–45 degrees or more. In addition, existing detectors typically handle them implicitly, producing low-confidence or drifting predictions without distinguishing uncertainty from genuine occlusion~\cite{burgos2013robust}.


Motivated by these challenges, we propose \textbf{OccFace}, an occlusion-aware landmark detection framework for universal human-like faces. OccFace builds on a heatmap-based backbone and augments it with (i) geometric evidence maps that help reweight heatmap responses for stable localization, and (ii) an occlusion prediction module that jointly estimates landmark coordinates and per-point visibility. The visibility head leverages complementary cues: local evidence around each landmark to address external occluders, and cross-landmark context to capture correlated self-occlusion under large pose. To supervise visibility, we combine manual occlusion labels with a landmark-aware masking strategy that generates synthetic occlusions and derives pseudo visibility signals from mask--heatmap overlap.

Beyond the model, we introduce an occlusion-aware evaluation suite. In addition to standard normalized mean error (NME)~\cite{300W}, we report localization performance separately on visible and occluded landmarks (NME$_{\mathrm{vis}}$, NME$_{\mathrm{occ}}$), and evaluate visibility prediction using standard classification metrics (Occ AP, F1@0.5, ROC-AUC). To support training and analysis in this setting, we further create \textit{Genie-Face}, a universal human-like face dataset annotated with our 100-point layout and per-point visibility across diverse domains. Fig.~\ref{fig:teaser} provides a visual summary of the task setting and our OccFace ecosystem.
Our contributions are summarized as follows:
\begin{itemize}
    \item To the best of our knowledge, we are the first to study universal human-like facial landmark detection with per-point visibility under both external occlusion and rotation-induced self-occlusion.
    \item We propose OccFace, a unified 100-point detector that jointly predicts coordinates and visibility, designed to handle both independent external occluders and correlated self-occlusion patterns.
    \item We create an accompanying ecosystem including an occlusion-aware evaluation suite, an occlusion-aware training recipe, and a universal dataset with 100-point landmarks and visibility annotations.
    \item Extensive experiments show improved robustness under occlusion and large head rotations, yielding more reliable landmarks for downstream face-centric applications such as auto-rigging and avatar animation.
\end{itemize}

\section{Related Work}
\label{sec:related}

\subsection{2D Facial Landmark Detection}

Classical face alignment relied on hand-crafted features and cascaded regressors~\cite{wu2018survey,kazemi2014one,FusionViT}. Modern methods are dominated by deep models that predict per-landmark heatmaps or regress coordinates, benefiting from strong spatial inductive bias~\cite{bodini2019review}. Stacked hourglass~\cite{Hourglass} and HRNet~\cite{HRNet} remain widely used backbones, while boundary-aware designs~\cite{WFLW,EffiPerception} further improve robustness on standard benchmarks~\cite{300W,COFW,WFLW}. Recent work further improves robustness via auxiliary cues such as uncertainty/visibility likelihood~\cite{kumar2020luvli}, error-bias correction~\cite{ADNet}, and transformer-based designs to handle pose and partial occlusion~\cite{chiang2025orformer}. However, most pipelines remain human-centric in both layout design and evaluation, and often do not expose reliable per-point visibility for downstream use.

\subsection{Landmark Detection on Human-like Faces}

Landmark detection for \emph{human-like} characters, such as stylized avatars, game assets, anthropomorphic animals, robots, is less explored than for real faces. Existing datasets and methods are typically domain-specific and use inconsistent or sparse landmark schemas~\cite{stricker2018manga,cai2021caricature}. Recent studies on stylized characters also show limited transfer from human-trained models without dedicated annotations~\cite{cheng2024stylizedfacepoint}. This motivates a unified, denser layout and a model that generalizes across diverse human-like faces.


\subsection{Facial Landmark Occlusion and Visibility Prediction}

Occlusion robustness is commonly addressed by visibility-aware modeling and occlusion augmentation. Visibility-aware methods estimate per-point visibility to down-weight occluded landmarks or enforce shape constraints under occlusion~\cite{burgos2013robust,wu2015occluded,liu2016occluded,kumar2020luvli}, while masking augmentations improve invariance but are not landmark-conditioned~\cite{devries2017cutout,zhong2020randomerase,chen2020gridmask}. These gaps motivate our landmark-conditioned occlusion reasoning with explicit visibility prediction for human-like faces.

\section{Method}
\label{sec:method}

\begin{figure}[!t]
  \centering
  \includegraphics[width=\linewidth]{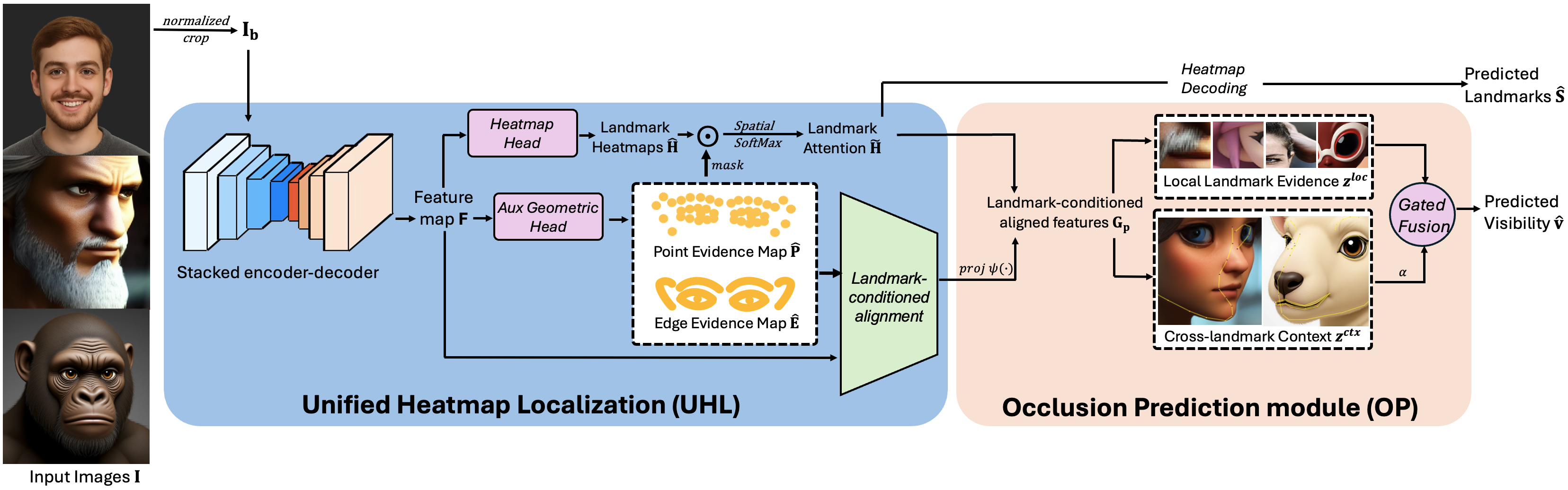}
  \caption{Overview of the OccFace evaluator. 
  We keep heatmap-based landmark localization and explicitly predict per-point visibility, enabling downstream modules to distinguish true occlusion from ambiguous predictions.
  }
  \label{fig:occface_overview}
\end{figure}

We propose \textbf{OccFace}, an occlusion-aware facial landmark detection framework for universal human-like faces that jointly predicts dense 2D landmark locations and per-point visibility.
Given an input image $\mathbf{I}$, OccFace outputs $P{=}100$ landmarks
$\hat{\mathbf{S}}=\{\hat{\mathbf s}_p\}_{p=1}^{P}$ with $\hat{\mathbf s}_p=(\hat{x}_p,\hat{y}_p)$,
and visibility scores $\hat{\mathbf v}=\{\hat{v}_p\}_{p=1}^{P}$.
Fig.~\ref{fig:occface_overview} summarizes the pipeline.
We first describe unified heatmap localization and the 100-point layout (Sec.~\ref{sec:heatmap}), then introduce visibility prediction (Sec.~\ref{sec:occ_module}), followed by occlusion-aware training/evaluation (Sec.~\ref{sec:training}) and our Genie-Face dataset (Sec.~\ref{sec:genie_face}).



\subsection{Unified Heatmap Localization for Human-like Faces}
\label{sec:heatmap}

Human-like faces exhibit large variation in geometry and rendering style, where direct coordinate regression can become limited under style changes and partial occlusion. We therefore adopt heatmap localization for stable, spatially grounded supervision.
This choice yields stable localization cues that are also essential for the visibility reasoning introduced in Sec.~\ref{sec:occ_module}.
Following the common face-alignment pipeline~\cite{SLPT}, we apply an affine crop-and-resize transform to obtain a normalized face crop $\mathbf{I}_b\in\mathbb{R}^{3\times h\times w}$ and perform learning in this coordinate system.

\begin{figure}[t]
  \centering
  \includegraphics[width=0.4\linewidth]{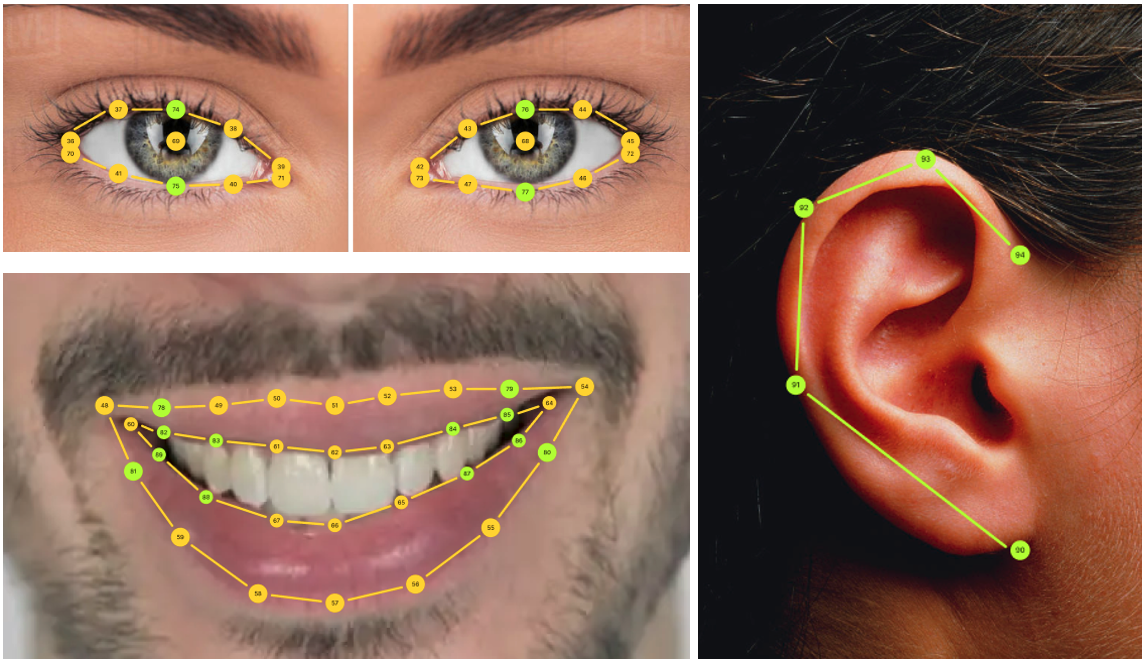}
  \caption{
  Our unified 100-point layout extends a standard 68-point schema~\cite{COCO} by adding (i) fine-grained eye anchors (incl.\ pupil/iris), (ii) inner-mouth structure, and (iii) explicit ear contours.
}
  \label{fig:layout_100pt}
\end{figure}

\mypara{Unified 100-point landmark layout.}
Conventional landmark layouts are human-centric and often under-specify expressive structures on stylized characters and non-human faces.
A clear example is the ear region in many mammals: its shape and posture can encode strong behavioral or affective cues, varying noticeably across states such as alertness or happiness. However, none of the existing human-centric dataset~\cite{300W,COFW,WFLW} contains annotation of the ear part.
We therefore introduce a unified 100-point layout (Fig.~\ref{fig:layout_100pt}) that preserves standard facial parts (brows/eyes/nose/mouth/jawline) while adding: (i) fine-grained eye points (pupil/iris), (ii) inner-mouth landmarks, and (iii) ear contours.
This denser layout improves cross-domain correspondences and provides richer visibility patterns (e.g., ears/inner-mouth) for occlusion supervision and evaluation.

\mypara{Stacked encoder--decoder backbone.}
We use a stacked hourglass-style encoder--decoder backbone~\cite{Hourglass,HRNet,3DifFusionDet} to predict landmark heatmaps and provide stable geometric cues that we later reuse for visibility reasoning.
We first apply a stem feature extractor (a small stack of convolutions and residual blocks) to map the normalized crop $\mathbf{I}_b$ to features $\mathbf{F}^{0}=f_{\mathrm{stem}}(\mathbf{I}_b)$, where $\mathbf{F}^{0}\in\mathbb{R}^{C\times h\times w}$ and $C$ denotes the feature-channel dimension.
We then apply $K$ stacked encoder--decoder modules for iteratively refinement. At stage $k$, the hourglass module $g_k(\cdot)$ takes the previous feature map $\mathbf{F}^{k-1}$ and produces refined features:
\begin{equation}
\mathbf{Z}^{k} = g_{k}(\mathbf{F}^{k-1}), \qquad k=1,\dots,K,
\end{equation}
where each $g_k(\cdot)$ follows an encoder--decoder structure with downsampling and upsampling, so that $\mathbf{Z}^{k}$ aggregates both local detail and broader context.
A projection layer $p_k(\cdot)$, which implemented as a small residual block followed by a $1{\times}1$ convolution, converts $\mathbf{Z}^{k}$ into the stage feature map:
$
\mathbf{F}^{k} = p_{k}(\mathbf{Z}^{k}).
$
From $\mathbf{F}^{k}$, a heatmap head $h_k(\cdot)$ predicts landmark response maps:
$
\hat{\mathbf{H}}^{k} = h^{k}(\mathbf{F}^{k}) \in \mathbb{R}^{P \times h \times w}.
$
We use the final heatmaps $\hat{\mathbf H}=\hat{\mathbf H}^{K}$ at inference. 
This stacked refinement lets later stages correct early landmark hypotheses using wider context while keeping precise spatial localization.

\mypara{Geometric maps for stable localization.}
To reduce drifting under ambiguous appearance, we predict two auxiliary geometric maps at the heatmap resolution. 
First, we group the $P$ landmarks into $N_E$ semantic edges, where each edge traces a meaningful facial boundary curve (e.g., jawline, brows, eyelids, lips, and ear contours).
This grouping provides long-range shape cues that help reduce drifting when local appearance is ambiguous.
Within the two auxiliary geometric maps, the \emph{point map} $\mathbf{P}\in\mathbb{R}^{P\times h\times w}$ places a small 2D Gaussian kernel around each landmark $\mathbf{s}_p$, and the \emph{edge map} $\mathbf{E}\in\mathbb{R}^{N_E\times h\times w}$ encodes semantic boundary curves as smooth distance-based heatmaps~\cite{chiang2025orformer}.
The backbone outputs $\hat{\mathbf P}$ and $\hat{\mathbf E}$ and we use them as additional geometric mask to reweight heatmaps before decoding.
For landmark $p$, let $\mathcal{E}(p)$ denote its associated semantic edge indices, we aggregate its associated edge evidence into a single map $\hat{\mathbf A}_p$ and form:
\begin{equation}
\hat{\mathbf A}_{p}(u,v)=\sum_{e\in\mathcal{E}(p)} \hat{\mathbf E}_{e}(u,v),
\qquad
\hat{\mathbf M}_{p}(u,v)=\hat{\mathbf P}_{p}(u,v)\cdot \hat{\mathbf A}_{p}(u,v).
\label{eq:edge_point_mask}
\end{equation}
We then apply spatial softmax on the reweighted heatmap and decode coordinates by expectation:
\begin{equation}
\tilde{\mathbf H}_p(u,v)=\mathrm{Softmax}_{u,v}\!\Big(\hat{\mathbf H}_p(u,v)\cdot \hat{\mathbf M}_p(u,v)\Big),\qquad
\hat{\mathbf s}_p=\sum_{u,v}(u,v)\,\tilde{\mathbf H}_p(u,v).
\label{eq:reweighted_decode}
\end{equation}
We will reuse $\hat{\mathbf P}$ and $\hat{\mathbf E}$ as landmark-conditioned inputs to the occlusion module (Sec.~\ref{sec:occ_module}); their supervision losses are included in Sec.~\ref{sec:training}.

\subsection{Occlusion Prediction Module for Per-Point Visibility}
\label{sec:occ_module}


Besides landmark coordinates, OccFace predicts an explicit visibility probability for each landmark. This is important for downstream use tasks such as tracking, animation, and correspondence, where we need to know whether a landmark is truly not visible (occluded) or simply hard to localize.
Occlusion in human-like faces mainly comes from two sources. \textit{External occlusion} (e.g., hair, hands, glasses, and accessories) is often local and may affect landmarks independently. In contrast, \textit{self-occlusion} caused by head rotation is more structured: landmarks on the far side (e.g., far eye, brow, cheek, and ear) often become invisible together. A purely local predictor may miss such correlated changes, while a purely global predictor may over-couple landmarks and hurt cases with small, independent occluders.
We therefore design the occlusion prediction module that combines a local branch with a cross-landmark context branch.



Let $\hat{\mathbf S}=\{\hat{\mathbf s}_p\}_{p=1}^{P}$ be the predicted landmark coordinates, where $\hat{s}_p=(\hat{x}_p,\hat{y}_p)$ and $P{=}100$. We predict a visibility probability $\hat{\mathbf{v}}=\{\hat{v}_p\}_{p=1}^{P}$, where $\hat{v}_p\in[0,1]$ indicates how likely landmark $p$ is visible in the image.
Given the normalized crop $\mathbf I_b$, the backbone provides features $\mathbf F\in\mathbb{R}^{C\times h\times w}$, landmark heatmaps $\hat{\mathbf H}$, and geometric maps $\hat{\mathbf P},\hat{\mathbf E}$ (Sec.~\ref{sec:heatmap}).
To obtain landmark-aligned cues for visibility, we reuse the soft attention map $\tilde{\mathbf H}_p$ and form per-landmark feature maps:
\begin{equation}
\mathbf{G}_p(u,v)=\tilde{\mathbf{H}}_p(u,v)\cdot \psi\!\left(\big[\,\mathbf{F}(u,v),\ \hat{\mathbf{P}}(u,v),\ \hat{\mathbf{E}}(u,v)\,\big]\right),
\label{eq:landmark_aligned_feat}
\end{equation}
where $\psi(\cdot)$ is a lightweight $1{\times}1$ projection that maps the concatenated cues into a common channel space. This yields a per-landmark representation that is spatially aligned with landmark $p$ and can be computed for all $P$ landmarks in a fully convolutional manner. It also keeps visibility prediction aligned with the same localization cues (heatmap responses and geometric maps), rather than introducing a separate occlusion detector.


\mypara{Local branch.}
The local branch predicts visibility for each landmark from its own neighborhood. We implement it using grouped and depth-wise convolutions applied to $\{\mathbf{G}_p\}_{p=1}^{P}$, followed by global average pooling, producing one logit per landmark $\mathbf{z}^{\text{loc}}\in\mathbb{R}^{P}$. As computation is grouped by landmark, this branch preserves point-wise independence and works well for localized external occluders.

\mypara{Cross-landmark context branch.}
To model rotation-driven self-occlusion, we add a context branch that allows information sharing across landmarks. We mix landmark channels before pooling (e.g., using a $1{\times}1$ convolution across the landmark dimension), so each landmark can use signals from other landmarks. This produces a second set of logits $\mathbf{z}^{\text{ctx}}\in\mathbb{R}^{P}$, which captures co-occlusion patterns such as multiple far-side 
landmarks becoming invisible together.

\mypara{Gated fusion and final visibility.}
We combine the two branches with a learnable gate that controls how much cross-landmark coupling is used.
Intuitively, the model can rely more on local evidence for independent external occlusion and increase coupling when rotation-driven self-occlusion is present.
We compute the final visibility logits as:
\begin{equation}
\mathbf{z} = \mathbf{z}^{\text{loc}} + \alpha \, \mathbf{z}^{\text{ctx}},
\end{equation}
where $\alpha$ is a per-landmark vector initialized close to 0, so that the module starts from the local-only behavior and gradually learns when cross-landmark context is helpful.
We then convert logits into probabilities by sigmoid function:
$
\hat{\mathbf{v}} = \sigma(\mathbf{z}).
$

This visibility head has two practical benefits.
First, it provides explicit per-point visibility that downstream modules can directly use, such as filtering or down-weighting occluded landmarks.
Second, the gate design avoids a fixed amount of landmark coupling: the model can learn when cross-landmark context is helpful. This helps keep robustness to external occlusion while still modeling structured self-occlusion.

\subsection{The Occlusion-Aware Metrics \& Training}
\label{sec:training}

This section summarizes (i) an occlusion-aware evaluation suite for assessing landmark localization and per-point visibility, and (ii) the training objectives.



\mypara{Localization metrics.}
Given redicted landmarks $\hat{\mathbf S}=\{\hat{\mathbf s}_p\}_{p=1}^{P}$ and ground truth $\mathbf S=\{\mathbf s_p\}_{p=1}^{P}$, we report normalized mean error:
$
\mathrm{NME}=\frac{1}{P}\sum_{p=1}^{P}\frac{\lVert \hat{\mathbf s}_p-\mathbf s_p\rVert_2}{d},
$
where $d$ is the dataset-specific normalization factor.
With ground-truth visibility $\mathbf v\in\{0,1\}^{P}$, we further split NME into visible and occluded subsets:
\begin{equation}
\mathrm{NME}_{\mathrm{vis}}=\frac{1}{|\mathcal V|}\sum_{p\in\mathcal V}\frac{\lVert \hat{\mathbf s}_p-\mathbf s_p\rVert_2}{d},
\qquad
\mathrm{NME}_{\mathrm{occ}}=\frac{1}{|\mathcal O|}\sum_{p\in\mathcal O}\frac{\lVert \hat{\mathbf s}_p-\mathbf s_p\rVert_2}{d},
\end{equation}
where $\mathcal V=\{p\,|\,v_p=1\}$ and $\mathcal O=\{p\,|\,v_p=0\}$.
This split shows whether errors mainly come from occluded regions or also affect visible landmarks.



\mypara{Visibility prediction metrics.}
For visibility probabilities $\hat{v}_p\in[0,1]$, we report Occ AP (precision--recall), F1 at threshold $\tau$ (we use $\tau{=}0.5$ unless stated), and ROC-AUC, aggregated over all landmarks 
.

\mypara{Training losses.}
For each training image, the model predicts heatmaps $\{\hat{\mathbf H}^{k}\}_{k=1}^{K}$, geometric maps $\hat{\mathbf P},\hat{\mathbf E}$, and visibility $\hat{\mathbf v}$.
We supervise landmarks with Gaussian heatmap targets 
$
\mathbf{H}_p(u,v)=\exp\!\left(-\frac{\lVert (u,v)-\mathbf{s}_p\rVert_2^2}{2\sigma^2}\right),
$
where $\sigma$ controls the peak width. We supervise all refinement stages with an MSE loss:
\begin{equation}
\mathcal{L}_{\mathrm{hm}}
=\sum_{k=1}^{K}\lambda_k \cdot \frac{1}{P}\sum_{p=1}^{P}\left\lVert \hat{\mathbf{H}}^{k}_{p}-\mathbf{H}_{p}\right\rVert_2^2,
\label{eq:hm_loss}
\end{equation}
where $\lambda_k$ weights each stage.
We additionally supervise the point/edge maps with MSE following~\cite{chiang2025orformer}:
$
\mathcal{L}_{\mathrm{pt}}=\frac{1}{P}\sum_{p=1}^{P}\left\lVert \hat{\mathbf{P}}_p-\mathbf{P}_p \right\rVert_2^2,
\mathcal{L}_{\mathrm{edge}}=\frac{1}{N_E}\sum_{j=1}^{N_E}\left\lVert \hat{\mathbf{E}}_j-\mathbf{E}_j \right\rVert_2^2.
$
These auxiliary objectives encourage sharp landmark-centered responses and consistent long-range boundary structure, improving stability when local appearance is ambiguous.
We train visibility with binary cross-entropy:
\begin{equation}
\mathcal{L}_{\mathrm{vis}}=\frac{1}{P}\sum_{p=1}^{P}\mathrm{BCE}(\hat{v}_p,v_p).
\label{eq:vis_loss}
\end{equation}

\mypara{Landmark-aware masking.}
Real occlusion annotations can be limited and biased toward visible landmarks.
We therefore augment visibility supervision with landmark-aware masking.
For a training image, we sample a random mask $\mathbf M\in\{0,1\}^{h\times w}$ and create a synthetically occluded view.
We derive pseudo visibility labels by measuring overlap between the mask and each landmark's ground-truth heatmap:
\begin{equation}
\tilde{v}_p=\mathbbm{1}\Big[\langle \mathbf M,\mathbf H_p\rangle<\delta\Big],
\end{equation}
where $\langle \cdot,\cdot\rangle$ denotes elementwise inner product and $\delta$ is a threshold.
Intuitively, if the mask covers the heatmap mass for landmark $p$, the landmark is labeled as occluded.
We then compute an auxiliary visibility loss $\tilde{\mathcal L}_{\mathrm{vis}}$ on masked samples, defined analogously to Eq.~\ref{eq:vis_loss} but with the pseudo labels $\tilde{v}$.


\mypara{Warm-start and final objective.}
Visibility prediction relies on landmark-conditioned features derived from predicted heatmaps and geometric maps, which become more reliable once heatmap responses form stable peaks. We therefore use a warm-start schedule: we first train the localization backbone with $\mathcal{L}_{\mathrm{hm}}$, $\mathcal{L}_{\mathrm{pt}}$, and $\mathcal{L}_{\mathrm{edge}}$, and then enable the visibility head and jointly optimize all losses. This schedule keeps localization stable and prevents the visibility head from learning from noisy early landmark hypotheses. In sum, the full objective is:
\begin{equation}
\mathcal{L}
=\mathcal{L}_{\mathrm{hm}}
+\lambda_{\mathrm{pt}}\mathcal{L}_{\mathrm{pt}}
+\lambda_{\mathrm{edge}}\mathcal{L}_{\mathrm{edge}}
+\lambda_{\mathrm{vis}}\mathcal{L}_{\mathrm{vis}}
+\lambda_{\mathrm{syn}}\tilde{\mathcal{L}}_{\mathrm{vis}},
\end{equation}
where $\lambda_{\mathrm{pt}},\lambda_{\mathrm{edge}},\lambda_{\mathrm{vis}},\lambda_{\mathrm{syn}}$ balance the contributions of each term.

\subsection{The Genie-Face Universal Human-like Face Dataset}
\label{sec:genie_face}

\begin{table}[t]
  \centering
  \small
  \setlength{\tabcolsep}{6pt}
  \resizebox{0.55\linewidth}{!}{
  \begin{tabular}{lrr}
    \toprule
    Domain & \# Images & Ratio (\%) \\
    \midrule
    Human-styled avatars  & 5,090 & 32.9 \\
    Rendered human faces  & 3,045 & 19.7 \\
    Real human photos & 2,003 & 13.0 \\
    Mammal faces & 3,783 & 24.4 \\
    Game character faces & 712 & 4.6 \\
    Alien-like faces & 511 & 3.3 \\
    Robot-like faces & 331 & 2.1 \\
    \midrule
    Total & 15,475 & 100.0 \\
    \bottomrule
  \end{tabular}
  }
  \caption{Genie-Face dataset composition.}
  \label{tab:data_sta}
\end{table}

\begin{figure}[!t]
  \centering
  \includegraphics[width=\linewidth]{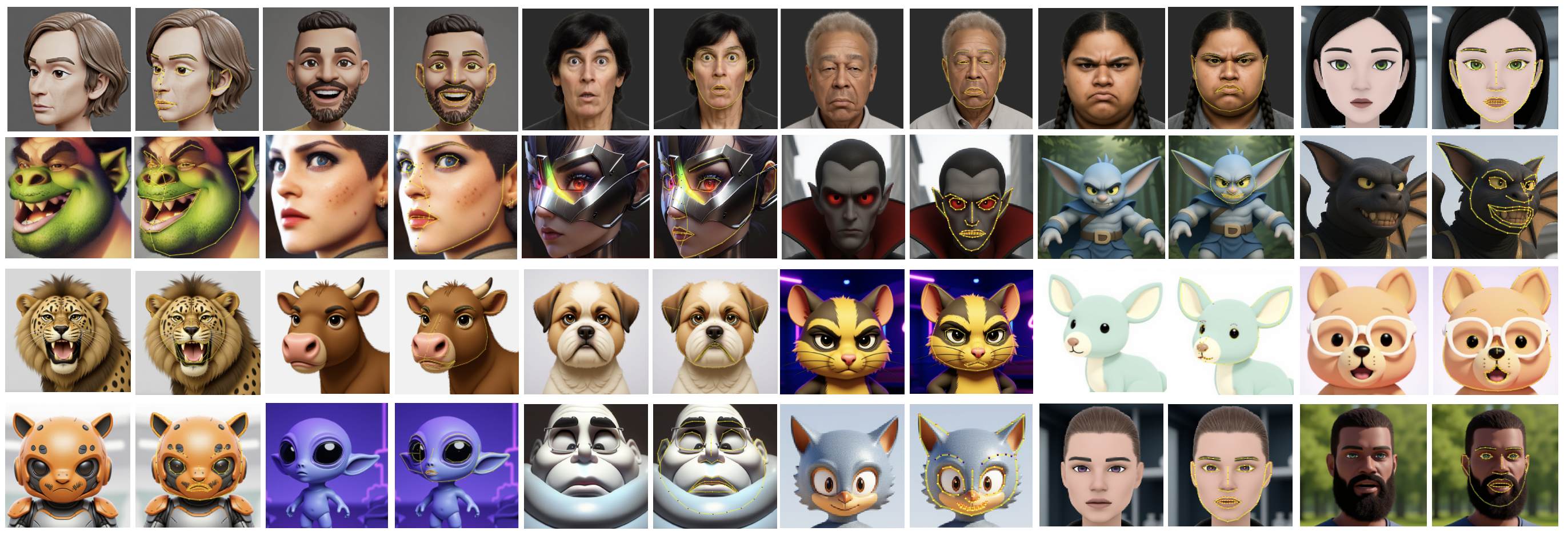}
  \caption{Samples from the Genie-Face dataset with annotation across diverse human-like domains. Landmarks are color-coded by visibility: yellow for visible and blue for invisible (occluded) points.}
  \label{fig:data_eg}
\end{figure}

Existing facial-landmark benchmarks are largely human-centric, with limited coverage of stylized or non-human geometry and strong self-occlusion under large head rotations.
To support occlusion-aware analysis in this broader setting, we create \textbf{Genie-Face}, a dataset of \emph{human-like faces} annotated with our unified 100-point layout and per-point visibility.

\mypara{Scope and sources.}
Genie-Face spans diverse facial structures and rendering styles under a consistent landmark definition, including real human photos, rendered human faces, stylized avatars/cartoon characters, and non-human human-like designs.
Table~\ref{tab:data_sta} list the dataset composition, while Fig.~\ref{fig:data_eg} show some samples of the dataset.
This diversity introduces large variation in proportions, heading orientation, materials (such as skin/fur/synthetic), occlusion, background, and parts that are uncommon in standard human layouts.

\mypara{Data collection.}
We aggregate images from multiple sources to increase diversity while following the same annotation rules.
In addition to carefully chosen real and rendered assets, we follow the classic synthetic data collection pipeline~\cite{AIGVE_Tool,ADGVE,liu2024surveyaigve} to include prompt-driven generated images to control viewpoint and head rotation, expression, and common occludes, with both clean and complex backgrounds.
A subset of candidate sets used in data collection is provided in the supplementary material.


\mypara{Annotation schema.}
Each image is annotated with our 100-point layout (Fig.~\ref{fig:layout_100pt}).
For every landmark, annotators additionally assign a binary visibility label indicating whether the landmark is visually observable.
Visibility accounts for both external occlusion and rotation-induced self-occlusion, where far-side structures may be hidden by the face itself.


\mypara{Annotation protocol and quality control.}
A landmark is labeled \emph{visible} if its semantic location can be identified on the image surface, and \emph{occluded} otherwise, such as covered by an object, self-occluded, or unobservable under extreme viewpoint.
We perform consistency checks and targeted review on difficult cases.


\section{Experiments}
\label{sec:exp}

\subsection{Facial Landmark Performance results}

\begin{table}[t]
  \centering
  \setlength{\tabcolsep}{4pt}
  \renewcommand{\arraystretch}{1.15}
  \begin{tabular}{l c c c c c c c}
    \toprule
    \multirow{2}{*}{Method} &
    \multicolumn{1}{c}{COFW} &
    \multicolumn{3}{c}{300W (NME$\downarrow$)} &
    \multicolumn{3}{c}{WFLW-Full} \\
    \cmidrule(lr){2-2}\cmidrule(lr){3-5}\cmidrule(lr){6-8}
    & NME$\downarrow$ & Full & Comm. & Chal. & NME$\downarrow$ & FR$\downarrow$ & AUC$\uparrow$ \\
    \midrule
    AWing~\cite{AWing}                 & 4.94 & 3.07 & 2.72 & 4.52 & 4.36 & 2.84 & 0.572 \\
    LUVLi~\cite{kumar2020luvli}        & --   & 3.23 & 2.76 & 5.16 & 4.37 & 3.12 & 0.577 \\
    ViTPose~\cite{vitpose}             & 4.85 & 3.02 & 2.63 & 4.34 & 4.26 & 2.94 & 0.588 \\
    ADNet~\cite{ADNet}                 & 4.68 & 2.93 & 2.53 & 4.58 & 4.14 & 2.72 & 0.602 \\
    HIH~\cite{HIH}                     & 4.63 & 3.09 & 2.65 & 4.89 & 4.08 & 2.60 & 0.605 \\
    SLPT~\cite{SLPT}                   & 4.79 & 3.17 & 2.75 & 4.90 & 4.14 & 2.76 & 0.595 \\
    STAR~\cite{STAR}                   & 4.62 & 2.90 & 2.52 & 4.46 & 4.03 & 2.32 & 0.611 \\
    ORFormer~\cite{chiang2025orformer} & \textbf{4.46} & 2.90 & 2.53 & 4.43 & 3.86 & 1.76 & 0.622 \\
    \textbf{OccFace (Ours)}            & 4.53 & \textbf{2.83} & \textbf{2.33} & \textbf{3.36} & \textbf{3.77} & \textbf{1.68} & \textbf{0.639} \\
    \bottomrule
  \end{tabular}
  \caption{Main landmark localization results on standard benchmarks. We report NME on COFW and 300W (lower is better). On WFLW-Full, we report NME, failure rate (FR; lower is better), and AUC (higher is better).}
  \label{tab:main_benchmarks_extended}
\end{table}

\begin{table}[t]
  \centering
  \setlength{\tabcolsep}{4pt}
  \renewcommand{\arraystretch}{1.15}
  \begin{tabular}{l c c c c c c}
    \toprule
    \multirow{2}{*}{Method} &
    \multicolumn{3}{c}{Genie-Face-74} &
    \multicolumn{3}{c}{Genie-Face-100}  \\
    \cmidrule(lr){2-4}\cmidrule(lr){5-7}
    & NME$\downarrow$ & FR$\downarrow$ & AUC$\uparrow$ &
      NME$\downarrow$ & FR$\downarrow$ & AUC$\uparrow$ \\
    \midrule
    ViTPose~\cite{vitpose}              & 5.36 & 3.06 & 0.437 & 5.89 & 3.32 & 0.401 \\
    ORFormer~\cite{chiang2025orformer}  & 4.78 & 2.85 & 0.583 & 5.32 & 2.99 & 0.432 \\
    \textbf{OccFace (Ours)}             & \textbf{2.87} & \textbf{1.44} & \textbf{0.716} & \textbf{3.01} & \textbf{1.57} & \textbf{0.685} \\
    \bottomrule
  \end{tabular}
  \caption{Occlusion-aware landmark localization results on Genie-Face. Columns are grouped by dataset (top header) and metrics (second header).}
  \label{tab:genie_face_benchmarks}
\end{table}

\begin{table}[t]
  \centering
  \setlength{\tabcolsep}{5pt}
  \renewcommand{\arraystretch}{1.15}
  \begin{tabular}{lcccccc}
    \toprule
    Method & NME$\downarrow$ & NME$_{\mathrm{vis}}\downarrow$ & NME$_{\mathrm{occ}}\downarrow$ & Occ AP$\uparrow$ & F1@0.5$\uparrow$ & ROC-AUC$\uparrow$ \\
    \midrule 
    ViTPose~\cite{vitpose} & 5.89 & 5.32 & 7.41 & 0.502 & 0.400 & 0.879    \\
    ORFormer~\cite{chiang2025orformer} & 5.32 & 4.94 & 6.16 & 0.521 & 0.417 & 0.897\\
    \textbf{OccFace(Ours)} & \textbf{3.01} & \textbf{2.71} & \textbf{5.22} & \textbf{0.582} & \textbf{0.492} & \textbf{0.963} \\
    \bottomrule
  \end{tabular}
  \caption{Occlusion-aware localization and visibility prediction on Genie-Face-100. All methods report localization (NME, NME$_\mathrm{vis}$, NME$_\mathrm{occ}$), visibility metrics (Occ AP, F1, ROC-AUC) are reported when a method outputs per-point visibility. Lower is better for NME; higher is better for Occ AP, F1, and ROC-AUC.}
  \label{tab:occ_vis_merged}
\end{table}


\begin{figure}[t]
    \centering
    \includegraphics[width=\linewidth]{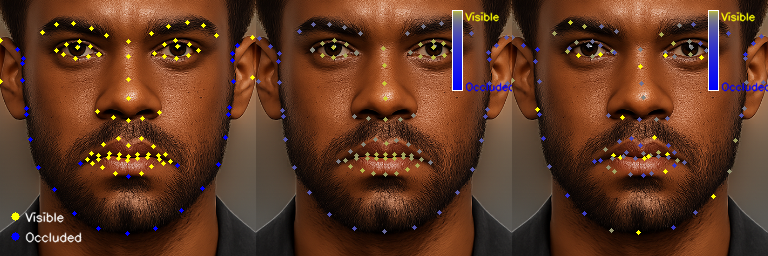}
    \includegraphics[width=\linewidth]{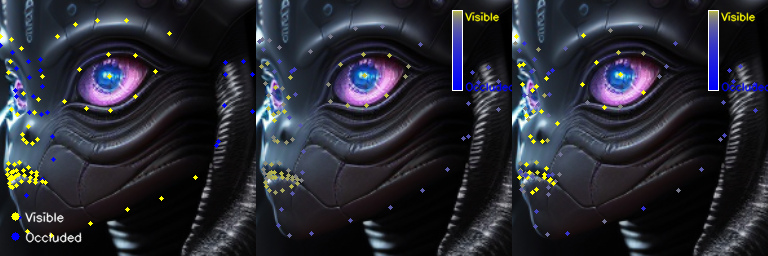}
    \caption{Qualitative comparison of landmark localization and visibility prediction on Genie-Face. Each triplet shows (left) ground truth, (middle) OccFace prediction, and (right) ORFormer prediction. Landmarks are color-coded by visibility: yellow for visible and blue for occluded. Since ORFormer does not natively predict visibility, we add a prediction head from its final feature space trained with our visibility objective. In addition to better landmark localization performance, OccFace produces more accurate visibility estimates, particularly for rotation-driven self-occlusion on far-side landmarks.}
    \label{fig:qualitative}
\end{figure}

\mypara{Experimental Setup.}
We evaluate OccFace on existing open facial landmark datasets~\cite{300W,COFW,WFLW} and our Genie-Face dataset, with a focus on human-like faces that contain frequent occlusion and large pose variations. We compare against strong baselines using the same input resolution and test-time pre-processing. Following each dataset's official metrics, we report normalized mean error (NME) for localization. For datasets with per-point visibility annotations, we additionally use our occlusion-aware suite (Sec.~\ref{sec:training}) to report $\mathrm{NME}_{\text{vis}}$, $\mathrm{NME}_{\text{occ}}$, and to evaluate visibility prediction with AP, recall, F1, and ROC-AUC.

\mypara{Landmark localization results.}
Tables~\ref{tab:main_benchmarks_extended} and~\ref{tab:genie_face_benchmarks} summarize the main localization results on existing open datasets and Genie-Face, respectively.
Overall, OccFace is competitive on existing human-centric facial landmark datasets and shows clearer advantages on harder cases with larger pose changes and heavier occlusion, which are central challenges in universal human-like faces.
On Genie-Face, the improvements are more noticeable.
OccFace performs significantly better than baselines, reducing NME from 43.78/58.32 (ORFormer) to 2.87/3.01.
This large gap indicates that Genie-Face poses significantly broader appearance and geometry variations than existing open facial landmark datasets. It also show that our unified layout and occlusion-aware design in OccFace generalize better to diverse human-like faces.

\mypara{Occlusion-aware analysis and visibility prediction.}
We further evaluate occlusion robustness on Genie-Face using our occlusion-aware metrics designed in Sec.~\ref{sec:training}.
Table~\ref{tab:occ_vis_merged}'s first three columns reports localization errors on visible versus occluded landmarks, showing that OccFace reduces error on occluded points while remaining accurate on visible points.
In addition, we report visibility prediction quality on Genie-Face-100 using Occ AP, F1, and ROC-AUC.
Shown in the last three columns, OccFace achieves the strongest visibility scores among the compared baselines, with clear gains in Occ AP and F1, indicating fewer missed occlusions and fewer false ``occluded'' predictions at a fixed threshold.
These visibility scores provide a practical signal for downstream systems to filter or down-weight landmarks that are likely occluded. 
Fig.~\ref{fig:qualitative} shows qualitative examples on both human and rendered human-like faces, illustrating that OccFace produces more reliable localization and visibility estimates.


\subsection{Ablations}

\begin{table}[t]
  \centering
  \setlength{\tabcolsep}{4pt}
  \renewcommand{\arraystretch}{1.15}
  \begin{tabular}{lcc}
    \toprule
    Variant & NME$\downarrow$ & NME$_{\mathrm{occ}}\downarrow$ \\
    \midrule
    \addlinespace[1pt]
    Heatmaps only           & 5.52 & 6.34 \\
    + point map             & 4.37 & 5.62 \\
    + point + edge maps (Ours)    & \textbf{3.01} & \textbf{5.22} \\
    \bottomrule
  \end{tabular}
  \caption{Ablation of auxiliary geometric maps on Genie-Face-100. We compare heatmap-only supervision with adding the point map and the edge map. We highlight NME$_{\mathrm{occ}}$ as it best reflects robustness under occlusion.}
  \label{tab:ablate_geomaps}
\end{table}

\begin{table}[t]
  \centering
  \setlength{\tabcolsep}{4pt}
  \renewcommand{\arraystretch}{1.15}
  \begin{tabular}{lcccc}
    \toprule
    Variant
    & NME$_{\mathrm{occ}}\downarrow$
    & Occ AP$\uparrow$
    & F1@0.5$\uparrow$
    & ROC-AUC$\uparrow$ \\
    \midrule
    Local-only branch ($z=z^{\mathrm{loc}}$)  & 6.87 & 0.493 & 0.396 & 0.883 \\
    Context-only branch ($z=z^{\mathrm{ctx}}$)& 6.48 & 0.524 & 0.405 & 0.892 \\
    Local + context (fixed sum, $\alpha{=}1$) & 5.42 & 0.553 & 0.455 & 0.921 \\
    \textbf{Local + context (learnable, ours)}    & \textbf{5.22} & \textbf{0.582} & \textbf{0.492} & \textbf{0.963} \\
    \bottomrule
  \end{tabular}
  \caption{Ablation of the visibility head design on Genie-Face-100. We compare local-only and context-only variants, and study how to combine both signals. 
  }
  \label{tab:ablate_vis_head}
\end{table}


\begin{table}[t]
  \centering
  \setlength{\tabcolsep}{4pt}
  \renewcommand{\arraystretch}{1.15}
  \begin{tabular}{lccccc}
    \toprule
    Training
    & NME$_{\mathrm{vis}}\downarrow$
    & NME$_{\mathrm{occ}}\downarrow$
    & Occ AP$\uparrow$
    & F1@0.5$\uparrow$
    & ROC-AUC$\uparrow$ \\
    \midrule
    w/o occAug  & 3.11 & 5.67 & 0.539 & 0.421 & 0.837 \\
    \textbf{+ occAug} & \textbf{2.71} & \textbf{5.22} & \textbf{0.582} & \textbf{0.492} & \textbf{0.963} \\
    \bottomrule
  \end{tabular}
  \caption{Ablation of \textbf{occAug} (landmark-aware masking with pseudo visibility labels) on Genie-Face-100. Adding occAug improves occluded-landmark localization and visibility prediction, while keeping NME$_{\mathrm{vis}}$ stable.}
  \label{tab:vis_ablate}
\end{table}

We conduct ablation studies to identify which components contribute most to OccFace's improvements under occlusion and large pose changes.
Unless otherwise specified, we keep the same training setup and report the occlusion-aware metrics from Sec.~\ref{sec:training}, with emphasis on $\mathrm{NME}_{\mathrm{occ}}$ and visibility metrics (Occ AP/F1/ROC-AUC), since they best reflect occlusion robustness.

\mypara{Effect of auxiliary geometric maps.}
Table~\ref{tab:ablate_geomaps} ablates the auxiliary geometric maps (introduced in Sec.~\ref{sec:heatmap}) used in our localization backbone.
As it shows, training with heatmaps only is less stable when local appearance is ambiguous, leading to larger errors on occluded landmarks and long curved structures.
Adding the point map improves landmark-centered supervision, and further adding the edge map provides boundary-aware shape cues.
Together, these maps reduce drifting and yield more reliable landmark-conditioned features that are later reused by the visibility head.

\mypara{Visibility head design: local/context branches and gated fusion.}
Table~\ref{tab:ablate_vis_head} compares variants of the visibility head (introduced in Sec.~\ref{sec:occ_module}) by changing the information branch path and the fusion rule.
Shown as a result, the local-only branch is effective for small, localized external occluders, but it is weaker under rotation-driven self-occlusion where multiple far-side landmarks disappear together.
The context-only branch captures these correlated patterns but can over-couple landmarks under localized occlusion.
Combining both branches achieves the best overall visibility quality, while keeping localization on visible landmarks stable.
In addition, the learnable gate further improves robustness compared to a fixed-weight sum by adapting the amount of cross-landmark coupling to the occlusion type.

\mypara{Landmark-aware masking for pseudo visibility.}
Table~\ref{tab:vis_ablate} evaluates landmark-aware masking and pseudo visibility supervision (introduced in Sec.~\ref{sec:training}).
As it demonstrates, adding occlusion augmentation improves visibility prediction metrics (Occ AP, F1, and ROC-AUC) and reduces NME$_{\mathrm{occ}}$, showing that training with manual visibility supervision only is limited by label sparsity and bias toward visible landmarks, while adding landmark-aware masking help the model learn more reliable occlusion cues.
Meanwhile, we find that this augmentation does not degrade NME$_{\mathrm{vis}}$, suggesting the model does not simply become conservative and mark hard-but-visible landmarks as occluded.


\subsection{Downstream Application: Avatar Animation}

\begin{figure}[t]
    \centering
    \includegraphics[width=\linewidth]{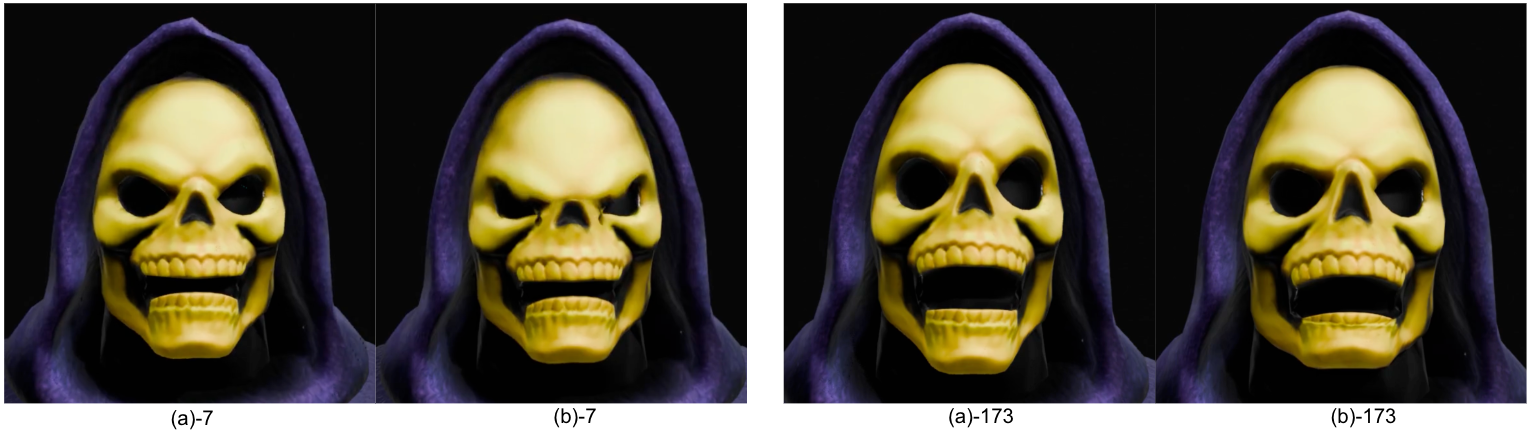}
    \caption{Avatar animation driven by landmarks from (a): OccFace and (b): ORFormer. We show frames 7 and 173 from a 216-frame animation video. OccFace's visibility-aware predictions yield more smoother and more consistent results.}
    \label{fig:app}
\end{figure}

To demonstrate the practical value of per-point visibility, we integrate OccFace into an avatar animation pipeline. 
Given a static 2D face image, we use predicted landmarks and visibility to drive facial animation on a 3D avatar. 
Fig.~\ref{fig:app} compares one sample animation result using OccFace versus ORFormer at two specific frames. It shows that OccFace could produce more stable facial animation with less jitter.



\section{Conclusion}
\label{sec:conclusion}


OccFace tackles universal human-like facial landmark detection under occlusion and large pose changes by jointly predicting 100-point landmarks and per-point visibility. With landmark-aware masking supervision, an occlusion-aware evaluation suite and the Genie-Face dataset, OccFace improves robustness on occluded landmarks while preserving accuracy on visible points, and provides visibility signals useful for filtering unreliable landmarks in downstream applications.

%
%
%

%
%
%
\bibliographystyle{splncs04}
\bibliography{main}
%





\clearpage
\setcounter{page}{1}
\maketitlesupplementary

\section{Candidate set taxonomy.}
To make the data generation/collection space explicit, we list representative candidate sets used in our dataset construction.
The taxonomy spans four orthogonal axes: (i) \emph{identity types} (e.g., human-like vs.\ non-human-like domains), (ii) \emph{appearance styles} (e.g., photorealistic vs.\ stylized renderings), (iii) \emph{scenes} (e.g., indoor/outdoor and background variations), and (iv) \emph{viewpoints} (e.g., yaw/pitch ranges).
Table~\ref{tab:data_topic} reports a subset of these candidates for reference; our full data collection additionally combines them with occlusion factors described in the main paper.

\begin{table}[ht]
  \centering
  \small
  \setlength{\tabcolsep}{5pt}
  \resizebox{0.8\linewidth}{!}{
  \begin{tabular}{l|p{0.70\linewidth}}
    \toprule
    Domain & Example candidates covered in Genie-Face (subset) \\
    \midrule
    Human-styled avatars
      & Outfit styles such as tactical outfit, desert explorer, cyberpunk jacket, medieval robe, business suit, and leather jacket; scenes such as modern laboratory, spaceship corridor, neon street at night, and forest clearing; controlled head turns (e.g., 15$^\circ$, 30$^\circ$, 45$^\circ$ left/right). \\
    \midrule
    Game characters
      & Role and style types such as warrior, mage, knight, ranger, ninja, assassin, fighter, soldier, pilot, hero; also non-human types such as alien, monster, beast, dragon, mech, and robot-like characters. \\
    \midrule
    Robotic characters
      & Robot identity types such as android, cyborg, automaton, mech unit, steel warrior, metal guardian, drone entity, synthetic being, and mechanical creature; with varied scenes such as laboratory, hangar, space station, stadium, and city streets. \\
    \midrule
    Animal characters
      & Mammal species including dog, cat, horse, cow, goat, sheep, pig, rabbit, raccoon, fox, wolf, tiger, lion, leopard, cheetah, giraffe, zebra, koala, kangaroo, panda, chimpanzee, and rhinoceros. \\
    \bottomrule
  \end{tabular}
  }
  \caption{A subset of candidate sets covered by Genie-Face across different human-like domains. These candidates are used during data collection and synthesis to ensure broad coverage beyond human faces.}
  \label{tab:data_topic}
\end{table}

\end{document}